\documentclass[10pt,twocolumn,letterpaper]{article}

\usepackage[pagenumbers]{iccv} 
\usepackage{multirow}
\usepackage{xcolor} 
\usepackage{amssymb} 
\definecolor{iccvblue}{rgb}{0.21,0.49,0.74}
\usepackage[pagebackref,breaklinks,colorlinks,allcolors=iccvblue]{hyperref}

\def\paperID{9236} 
\def\confName{ICCV}
\def\confYear{2025}

\title{MuseTalk: Real-Time High-Fidelity Video Dubbing via Spatio-Temporal Sampling}

\author{
Yue Zhang$^{1}$$^{*}$, Zhizhou Zhong$^{1}$$^{*}$, Minhao Liu$^{1}$\thanks{Equal contribution.}, Zhaokang Chen$^{1}$, Bin Wu$^{1}$\thanks{Corresponding author}, \\
Yubin Zeng$^{1}$, Chao Zhan$^{1,2}$\thanks{Work performed during an internship at Tencent Music Entertainment}, Yingjie He$^{1}$, Junxin Huan$^{1}$, Wenjiang Zhou$^{1}$ \\
\textsuperscript{1} Lyra Lab, Tencent Music Entertainment \\
\textsuperscript{2} The Chinese University of Hong Kong, Shenzhen
}

\begin{document}
\maketitle

\begin{abstract}
Real-time video dubbing that preserves identity consistency while achieving accurate lip synchronization remains a critical challenge. 
Existing approaches face a trilemma: diffusion-based methods achieve high visual fidelity but suffer from prohibitive computational costs, while GAN-based solutions sacrifice lip-sync accuracy or dental details for real-time performance. We present MuseTalk, a novel two-stage training framework that resolves this trade-off through latent space optimization and spatio-temporal data sampling strategy. 
Our key innovations include:
(1) During the Facial Abstract Pretraining stage, we propose Informative Frame Sampling to temporally align reference-source pose pairs, eliminating redundant feature interference while preserving identity cues.
(2) In the Lip-Sync Adversarial Finetuning stage, we employ Dynamic Margin Sampling to spatially select the most suitable lip-movement-promoting regions, balancing audio-visual synchronization and dental clarity.
(3) MuseTalk establishes an effective audio-visual feature fusion framework in the latent space, delivering 30 FPS output at 256×256 resolution on an NVIDIA V100 GPU.
Extensive experiments demonstrate that MuseTalk outperforms state-of-the-art methods in visual fidelity while achieving comparable lip-sync accuracy. 
The code is made available at \href{https://github.com/TMElyralab/MuseTalk}{https://github.com/TMElyralab/MuseTalk}
\end{abstract}

\section{Introduction}
\label{sec:intro}

\begin{figure}[tbp]
  \centering
   \includegraphics[width=\linewidth]{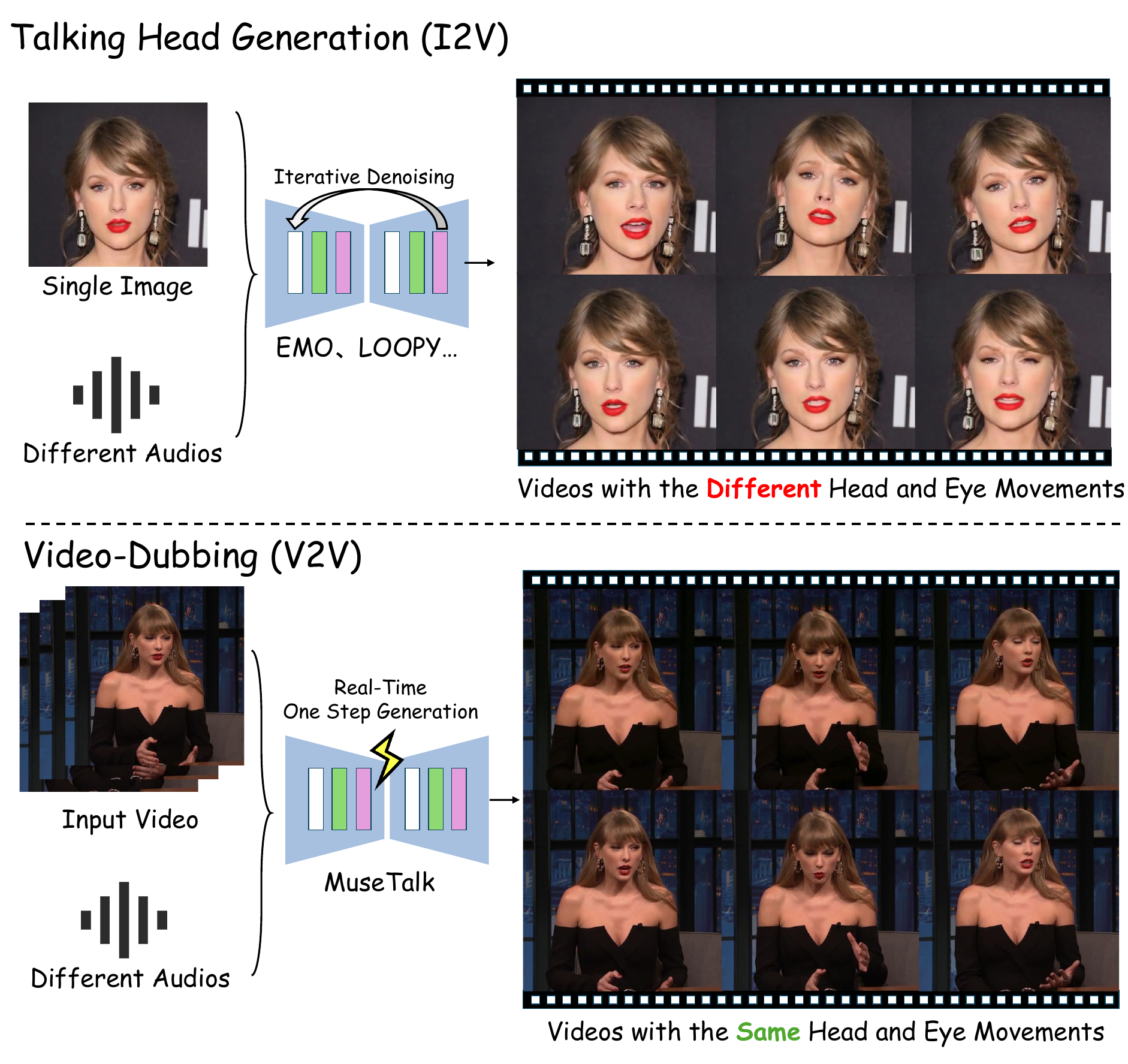}
   \caption{The difference between the talking head generation and the video dubbing. 
   Zoom in to see the differences in the lip area.
   MuseTalk can efficiently generate video frames in one step for video dubbing task.
   }
   \label{fig:fig_1}
\end{figure}

Virtual human generation~\cite{jia2024human,xing2024survey, xu2024mambatalk, tian2024emo, hu2024animate, guo2024liveportrait} is an important research field in computer vision. One significant application is generating lip movements that match the pronunciation of the target language for multilingual films or animations~\cite{prajwal2020lip}.It removes the mismatch between lip movements and speech in traditional dubbing, enhancing audience immersion without reshooting actors.

Recently, Image-to-Video (I2V) talking face generation methods~\cite{zhen2023human} have shown significant potential in creating virtual actors by generating highly realistic avatars for specific identities. 
One-shot talking face generation methods based on diffusion models~\cite{ho2020denoising, rombach2022high}, such as EMO~\cite{tian2024emo}, EchoMimic~\cite{chen2024echomimic}, and LOOPY~\cite{jiang2024loopy}, have gained popularity.
These methods allow users to create a video with good audio-visual consistency by uploading just a single image and an audio clip. 
However, their reliance on iterative denoising processes restricts their suitability for real-time applications. 
In the generated videos, the model autonomously adjusts characters' head and eye movements based on the audio, as illustrated in~\cref{fig:fig_1}.

In this paper, we investigate one-shot video-dubbing, a Video-to-Video (V2V) technique that focuses on preserving the original actor's head and eye movements while selectively modifying only the lip movements, without requiring additional model retraining.
Existing one-shot video-dubbing methods can be broadly categorized into two main paradigms: diffusion-based approaches \cite{li2024latentsync, ma2025sayanything} and GAN-based techniques \cite{zhang2023dinet, cheng2022videoretalking, prajwal2020lip, wang2023seeing}. 
While diffusion models have demonstrated remarkable capabilities for video-dubbing tasks, their reliance on extensive training data and multi-step inference processes makes them computationally expensive and impractical for local deployment by media professionals and AI artists.

Existing GAN-based methods~\cite{zhang2023dinet, cheng2022videoretalking, prajwal2020lip, wang2023seeing} often fall short in video quality, frequently producing blurry or distorted regions that negatively impact visual fidelity.
Additionally, these methods often fail to preserve the original actor's identity accurately, leading to noticeable changes in facial appearance during the dubbing process. 
Such issues significantly limit their practical applications. 
Despite well-known training instabilities in GANs~\cite{goodfellow2014generative, mescheder2018training}, their ability to generate outputs in a single step provides a promising solution to real-time application.
Driven by the computational efficiency and cost-effectiveness of GANs, we investigate novel one-shot lip-sync generation methods that achieve high-quality results while maintaining efficiency.

This paper introduces MuseTalk, a GAN-based real-time one-shot video-dubbing framework. 
Specifically, to reduce user costs, we design a one-step face generator in the VAE~\cite{kingma2013auto} latent space. 
MuseTalk addresses key training challenges in GAN-based video-dubbing through a carefully designed two-stage training process. 
A novel spatio-temporal sampling strategy is proposed to improve identity consistency and lip movement accuracy.
We first implement mild pretraining using latent space inpainting to enhance the model's ability for facial abstract prediction. 
During this stage, we introduce Informative Frame Sampling to select key frames.
Subsequently, we incorporate audio-visual synchornize loss and GAN loss, with the latter focusing on optimizing the mouth region. 
Here, Dynamic Margin Sampling is employed to spatially select critical facial regions that promote better lip movement learning.

In summary, our contributions are three-fold: 
(i) We propose MuseTalk, a GAN-based video-dubbing framework based on latent space inpainting, which enables real-time generation of high-fidelity lip-synced videos; 
(ii) We introduce a comprehensive two-stage training framework that resolves the conflict between GAN loss and audio-visual synchornize loss, achieving a balance between lip movement accuracy and teeth clarity; 
(iii) We propose a novel spatio-temporal sampling strategy. Specifically, we design Informative Frame Sampling at the frame level to bridge the gap between training and inference, and Dynamic Margin Sampling at the region level to promote lip movement learning in adversarial training. 
Extensive experimental results demonstrate the efficacy of MuseTalk, even compared to diffusion-based methods.
\section{Related Work}
\label{sec:related_work}

\subsection{Talking Head Generation}
\label{sec: rw_1}
In recent years, audio-driven talking head generation methods have attracted significant attention and provide valuable insight for video dubbing techniques. Talking head generation can be categorized into NeRF-based~\cite{peng2024synctalk, tang2022real, liu2022semantic, guo2021ad}, GAN-based~\cite{chen2019hierarchical, zhou2019talking, zhou2021pose, meshry2021learned, das2020speech}, and diffusion-based~\cite{tian2024emo, chen2024echomimic, xu2025vasa, jiang2024loopy, xu2024hallo, chencafe} approaches.

NeRF-based methods require identity-specific videos for training and additional rendering~\cite{mildenhall2021nerf} time, with early methods like AD-NeRF~\cite{guo2021ad} taking several seconds per frame. Despite the real-time rendering achieved by integrating Instant-NGP~\cite{muller2022instant, tang2022real}, retraining is needed for new identities. Diffusion-based methods, such as EMO Portrait~\cite{tian2024emo}, employ a two-stage training process by integrating ReferenceNet~\cite{hu2024animate}, temporal layers~\cite{guo2023animatediff}, and audio attention layers into Stable Diffusion models~\cite{rombach2022high}. Similar strategies are used in LOOPY~\cite{jiang2024loopy}, Hallo~\cite{xu2024hallo}, and EchoMimic~\cite{chen2024echomimic}. 
These methods allow for one-shot vivid talking head generation from images, but are computationally expensive. 

In contrast, GAN-based methods generate images in one step.
Early methods~\cite{chen2018lip, chen2019hierarchical, zhou2019talking} fail to maintain identity consistency and accurate lip movement. 
To address this, methods like MakeItTalk~\cite{zhou2020makelttalk} and SadTalker~\cite{zhang2023sadtalker} adopt multistage inference, separating audio-to-motion and motion-to-video modeling. 
Although this improves the results, it increases the computational overhead and complexity.

\begin{figure*}[tbp]
  \centering
   \includegraphics[width=\linewidth]{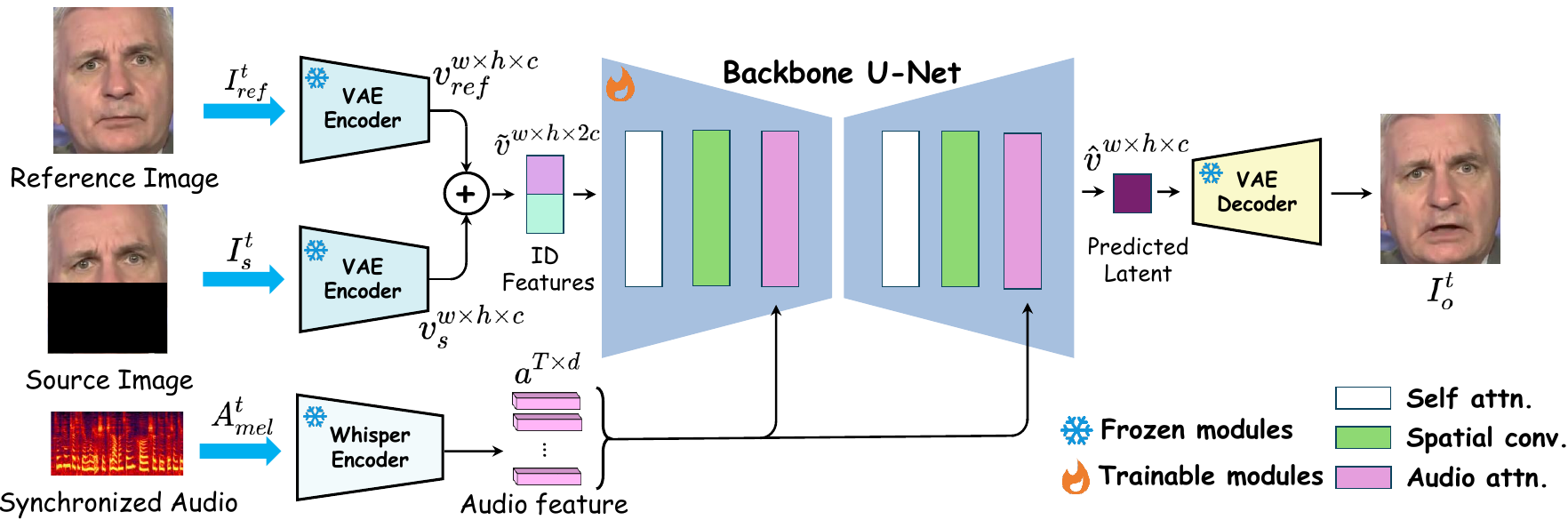}
   \caption{
   Illustration of MuseTalk's framework. 
   We first encode a reference facial image and an occluded lower half target image into perceptually equivalent latent space. 
   Subsequently, we employ a multimodal U-Net to effectively fuse audio and visual features at various scales. 
   Consequently, the decoded results from the latent space yield more realistic and lip-synced talking face visual content.}
   \label{fig:fig_pipeline}
\end{figure*}

\subsection{Video Dubbing}
\label{sec: rw_2}
Video dubbing focuses on replacing the mouth region of a source face based on driving audio. The most common approach~\cite{prajwal2020lip} involves using a mask in the lower half of the face or the lip region to guide the model to create new visual effects only in these areas, as shown in~\cref{fig:fig_pipeline}. 
Early methods~\cite{prajwal2020lip, zhang2023dinet, cheng2022videoretalking, wang2023seeing, guan2023stylesync, sun2022masked, zhong2023identity} predominantly relied on GANs. 
Although these methods achieved lip movements that were relatively consistent with the audio content, they often struggled with maintaining identity consistency and reconstructing clear teeth details. 
DI-Net~\cite{zhang2023dinet} attempted to train models on high-resolution data, which compromised lip accuracy. 
StyleSync~\cite{guan2023stylesync} highlighted this dilemma, noting that forcing the model to restore too many lip-relevant details might interfere with learning.

Recently, methods such as LatentSync~\cite{li2024latentsync} and DiffTalk~\cite{shen2023difftalk} have leveraged the strong detail generation capabilities of the Latent Diffusion paradigm~\cite{rombach2022high} for video dubbing tasks. 
LatentSync integrates the SyncNet loss~\cite{prajwal2020lip} to enhance lip motion-audio alignment during the one-step denoising process. 
Despite improvements in audio-visual synchornize and clarity, these methods suffer from the heavy inference burden.

\section{Method}
\label{sec:method}

\subsection{Overview}
\label{subsec:method_1}
We propose MuseTalk, a GAN-based one-step generation framework operating in the VAE latent space, building on insights from prior work. This section outlines the technical implementation and design principles of MuseTalk.

Through experimentation, we found that simultaneously optimizing the SyncNet loss~\cite{prajwal2020lip} and GAN loss in a single training stage for a randomly initialized model leads to training instability. We further discuss this issue in the supplementary materials. In contrast, MuseTalk introduces a novel two-stage training strategy to mitigate this problem.

The first stage, \textbf{Facial Abstraction Pretraining}, establishes foundational visual representations using our proposed \textbf{Informative Frame Sampling (IFS) }mechanism. In the second stage, \textbf{Lip-Sync Adversarial Finetuning}, we introduce \textbf{Dynamic Margin Sampling (DMS)} to balance adversarial training objectives with lip-synchronization constraints, enabling effective optimization of both aspects.

\subsection{Network Pipelines}
\label{subsec:method_2}
MuseTalk is designed to seamlessly integrate audio and visual information while maintaining efficient one-step inference capabilities. 
Unlike conventional GAN-based methods~\cite{prajwal2020lip, wang2023seeing}, which use independent encoders for audio and visual data, we leverage a multimodal U-Net architecture~\cite{ronneberger2015u} as the backbone of the generator. 
To further enhance computational efficiency, we refer to the Latent Diffusion approach~\cite{rombach2022high}, shifting the learning task from the pixel domain to the latent space.

\noindent \textbf{Identity Feature Handling.}
For video dubbing, the generated video must retain the identity of the original reference image, which necessitates integrating identity information into the network. 
Previous diffusion-based methods~\cite{tian2024emo, jiang2024loopy} achieve this by incorporating a ReferenceNet that adjusts each attention layer to inject identity information. 
While this approach is effective for multi-step denoising processes, it can be overly computationally expensive for one-step predictions.
Instead, we simplify the process by concatenating identity information along the channel dimension at the U-Net input.

Specifically, we pass the upper half of the source image at time \( t \) and the full-face reference image (captured at a different moment) through a VAE encoder.
The encoded features are then concatenated along the channel dimension to form a comprehensive image feature representation \(v^{w \times h \times 2c}\), where \(w\) and \(h\) denote the width and height of the feature, respectively. 
As illustrated in~\cref{fig:fig_pipeline}, an occluded lower half of the ground truth image \(I_{s}^{t}\) and a reference identity image \(I_{\text{ref}}^{t}\) at time \(t\) are each passed through the VAE encoder, producing outpus \(v_{\text{ref}}^{w \times h \times c}\) and \(v_{m}^{w \times h \times c}\). 
Experimental results in~\cref{tab:performance} demonstrate that this straightforward yet effective design provides robust identity control.

During inference,  only a single input frame at time \( t \)  is required. This frame is used as both \( I_{\text{ref}}^{t} \) and \( I_{s}^{t} \). 
The generated \( I_{o}^{t} \) is subsequently overlaid onto the original image using advanced face parsing and blending techniques. 
Detailed descriptions are provided in the supplementary materials. 
The construction of reference and source images during training is elaborated upon in subsequent sections.

\begin{figure}[tbp]
  \centering
   \includegraphics[width=\linewidth]{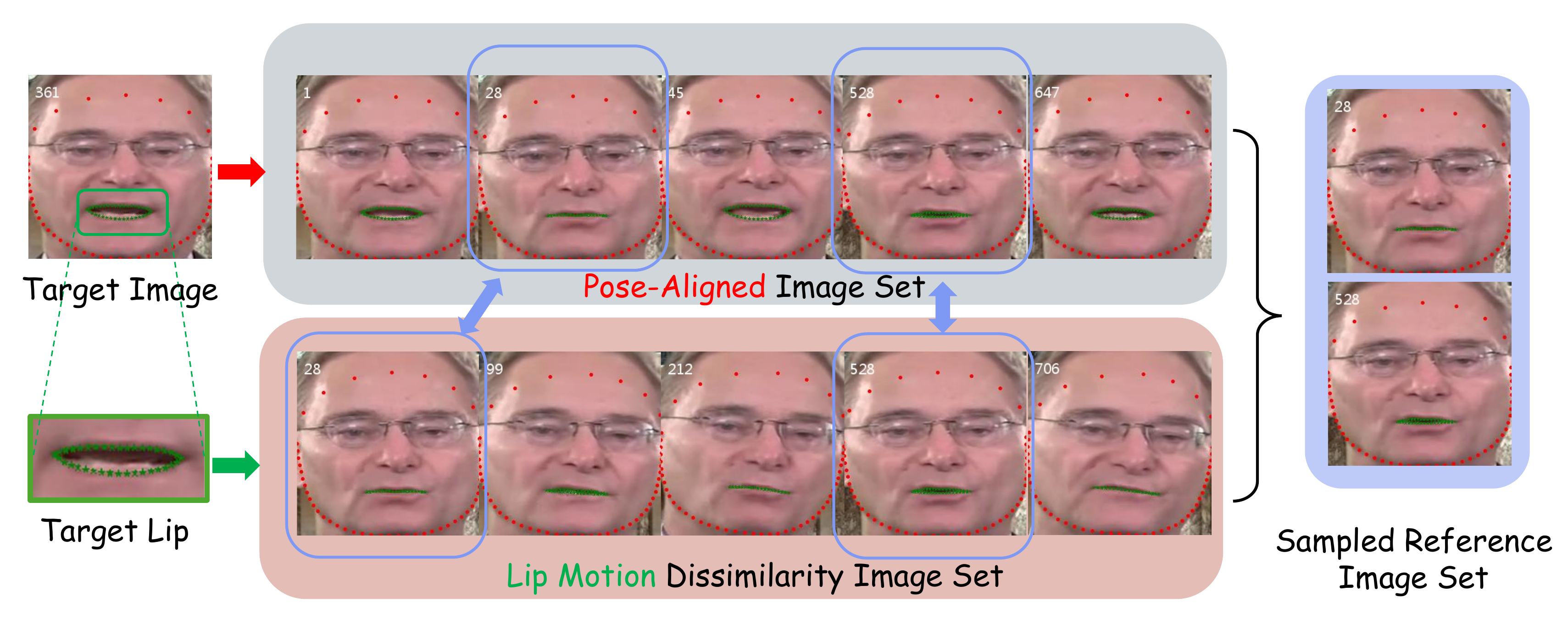}
   \caption{
   The illustration of proposed Informative Frame Sampling mechanism. 
   We calculate the pose and lip similarity based on Euclidean distance between facial landmarks.
   }
   \label{fig:fig_IFS}
\end{figure}

\noindent \textbf{Audio Feature  Handling.}
Following established practices~\cite{tian2024emo, jiang2024loopy, chen2024echomimic}, we utilize a pre-trained audio encoder~\cite{baevski2020wav2vec, radford2023robust} to extract audio features and inject them into the U-Net through its cross-attention layers. To optimize inference speed, we employ the lightweight Whisper-Tiny model~\cite{radford2023robust} to process an audio segment centered at time \(t\) with duration \(T\). 
The selected audio segment is first resampled to 16,000 Hz and converted into an 80-channel log-magnitude Mel spectrogram, denoted as \(A_{mel}^{t} \in \mathbb{R}^{T \times 80}\). 
The resulting audio feature has dimensions \(a^{T \times d}\), where \(d=384\).

\subsection{Facial Abstract Pretraining}
\label{subsec:method_3}
\noindent \textbf{Observation.} 
We observed that joint optimization of multiple losses during early training stages leads to unstable convergence, particularly when combined with adversarial objectives. 
To address this challenge, we adopt a phased training approach where the first stage focuses on cultivating facial abstract inpainting capabilities through mild optimization strategies.

\noindent \textbf{Loss Function.}
At this stage, we employ stable reconstruction losses instead of adversarial training objectives. To focus the model's attention as much as possible on the facial region, we crop the images along the edges of the face, thereby reducing the interference from background inpainting tasks on the model, as illustrated in~\cref{fig:fig_DMS}. Given a synthesized talking face image $I^{t}_{o}$ and its ground truth counterpart $I^{t}_{gt}$, we formulate the optimization target as:
\begin{equation}
\mathcal{L_{\text{stage1}}} = \left \| I^{t}_{o} - I^{t}_{gt} \right \|_1 + \lambda_{vgg} \left \| \mathcal{V}(I^{t}_{o}) - \mathcal{V}(I^{t}_{gt}) \right \|_2,
\label{loss:stage1}
\end{equation}
where $\mathcal{V}$ denotes the feature extractor of VGG19~\cite{simonyan2014very}. 
Utilizing solely the L1 loss tends to produce overly smoothed facial reconstructions. 
In contrast, perceptual loss~\cite{johnson2016perceptual} facilitates the learning of high-frequency visual patterns, particularly in capturing transitional facial features such as sideburn textures and incipient dental structures. 

\noindent \textbf{Informative Frame Sampling.} 
Previous GAN-based approaches~\cite{prajwal2020lip, zhang2023dinet, cheng2022videoretalking} rely on random sampling to obtain the reference image \(I^{t}_{ref}\). 
However, this method introduces a significant gap between training and inference phases. 
Specifically, during training, the reference image \(I^{t}_{\text{ref}}\) and the ground truth image \(I^{t}_{\text{gt}}\) often exhibit different head poses. In contrast, during inference, \(I^{t}_{\text{ref}}\) and \(I^{t}_{\text{gt}}\) share the same pose.
This discrepancy makes it challenging for models to generalize well across different scenarios. 
We introduce a novel Informative Frame Sampling (IFS) strategy to address the training-inference discrepancy by focusing the model on lip movement generation.
The IFS strategy aims to construct data pairs \(I^{t}_{ref}\) and \(I^{t}_{gt}\) that retain relevant texture details while filtering out redundant or distracting information. As illustrated in~\cref{fig:fig_IFS}, the process involves three key steps:

\begin{enumerate}
\item \textbf{Pose Alignment:} We calculate head pose similarity using chin landmark distances and select the most similar frames to form the Pose-Aligned Set \(\mathcal{E}_{\text{pose}}\). 

\item \textbf{Distinct Lip Movement:} We compute inner-lip landmark differences to identify frames with distinct lip movements, forming the Lip Motion Dissimilarity Set \(\mathcal{E}_{\text{mouth}}\). 

\item \textbf{Intersection Selection:} We choose the intersection \(\mathcal{E}_{\text{pose}} \cap \mathcal{E}_{\text{mouth}}\) as the Sampled Reference Image Set \(\mathcal{E}\), sort it by similarity, and select the top \(k\) subset as \(I^{t}_{\text{ref}}\). 
We later describe the optimal value of \(k\) in~\cref{ablation-IFS}. 
\end{enumerate}

\subsection{Lip-Sync Adversarial Finetuning}
\label{subsec:method_4}
\begin{figure}[tbp]
  \centering
   \includegraphics[width=\linewidth]{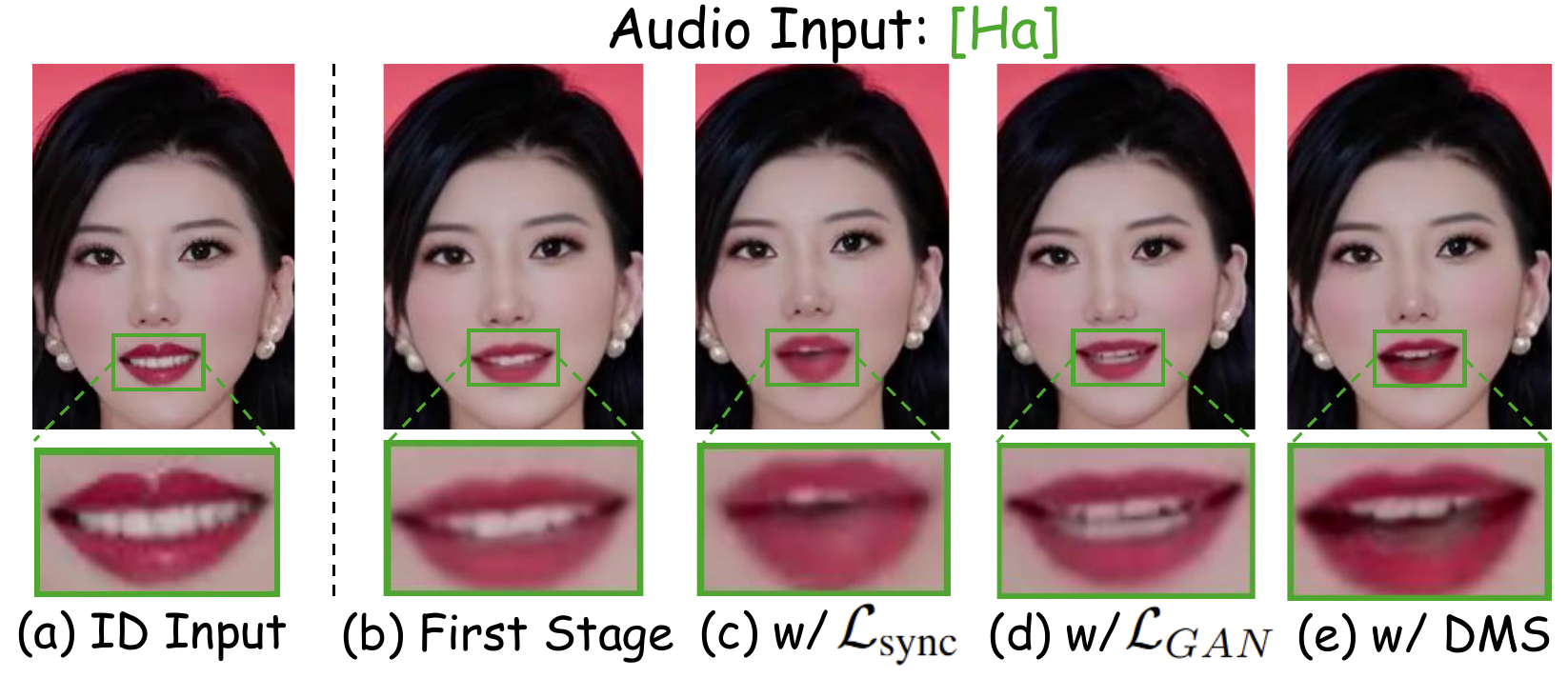}
   \caption{
    (a) Identity image during Inference.
    (b) First-stage model generates smooth teeth.
    (c) SyncNet loss promotes accurate lip movements but causes blurring.
    (d) GAN loss enhances clear teeth but replicates the original lip.
    (e) After applying DMS, both accurate lip movements and clear teeth are generated.
    }
   \label{fig:fig_loss_ab}
\end{figure}

\noindent \textbf{Optimization Objective.}
The primary goal of this stage is to enhance the model's capability in generating realistic dental details while ensuring precise lip movements. Building upon the initial training phase, where the model learns to extract facial abstract information from audio inputs, we observe that the outputs tend to exhibit overly smoothed teeth and replicated lip motions from reference images, as demonstrated in~\cref{fig:fig_loss_ab}(b). 
To overcome these limitations, we incorporate two loss functions: an adversarial loss~\cite{mao2017least} and a SyncNet loss~\cite{prajwal2020lip}.

\noindent \textbf{Adversarial Loss.}
The adversarial loss~\cite{gulrajani2017improved} is designed to enable the generator to capture intricate details by competing against two discriminators: one focused on the entire face \( \mathcal D_{face}\) and another specifically targeting the lip region \( \mathcal D_{lip}\). For the lip discriminator \( \mathcal D_{lip}\), the input region is carefully cropped based on the lip landmarks, \textbf{expand} to a fixed size of and fed into the network without any resizing operations. 
We chose the expand method over resizing because resizing degrades lip generation quality, resulting in inaccurate and unrealistic mouth shapes.
The optimization objective for the adversarial loss is formulated as:

\begin{equation}
\mathcal{L}_{adv} = \mathcal{L}_{adv,face}+\mathcal{L}_{adv,lip},
\end{equation}
where 
\begin{equation}
\mathcal{L}_{adv,face} = -\mathbb{E}_{A_{mel}^{t},I^{t}_{ref}} [ \mathcal D_{face}(I^{t}_{o})],
\end{equation}
\begin{equation}
\mathcal{L}_{adv,lip} = -\mathbb{E}_{A_{mel}^{t},I^{t}_{ref}} [ \mathcal D_{lip}(I^{t}_{lip})].
\end{equation}

\noindent \textbf{Sync Loss.}
The SyncNet loss promotes lip movement learning by aligning the generated frames with the audio~\cite{li2024latentsync, prajwal2020lip}. Specifically, we apply a SyncNet \(\mathcal S\) that takes \(N\) pairs of audio and image frames as input. The output features are then used to calculate the cosine similarity. The optimization objective is:
\begin{equation}
\mathcal{L}_{sync} = \frac{1}{N} \sum_{i}^{N} -\log[{CosSim}(\mathcal S(A^{i}_{mel}, I^{i}_o))].
\label{loss:syncloss}
\end{equation}

The overall optimization objective for this stage is:
\begin{align}
\mathcal{L}_{\text{stage2}} = \mathcal{L}_{\text{stage1}} + \lambda_{adv} \mathcal{L}_{adv} 
+ \lambda_{sync} \mathcal{L}_{\text{sync}}.
\label{loss:stage2}
\end{align}

\noindent \textbf{Dynamic Margin Sampling.}
When optimizing \(\mathcal{L}_{\text{stage2}}\), we observe conflicts between the adversarial loss and the SyncNet loss. 
Specifically, optimizing the adversarial loss alone can produce clear teeth but causes the model to replicate the reference lip movements, especially when the reference image shows teeth, as illustrated in~\cref{fig:fig_loss_ab}(d). Conversely, as shown in~\cref{fig:fig_loss_ab}(c), optimizing the SyncNet loss alone enables the model to close the mouth during silent periods but results in blurry lip movements when speaking. When both losses are optimized simultaneously, the SyncNet loss becomes difficult to converge, and the model's behavior tends towards that shown in~\cref{fig:fig_loss_ab}(d).

\begin{figure}[tbp]
  \centering
   \includegraphics[width=\linewidth]{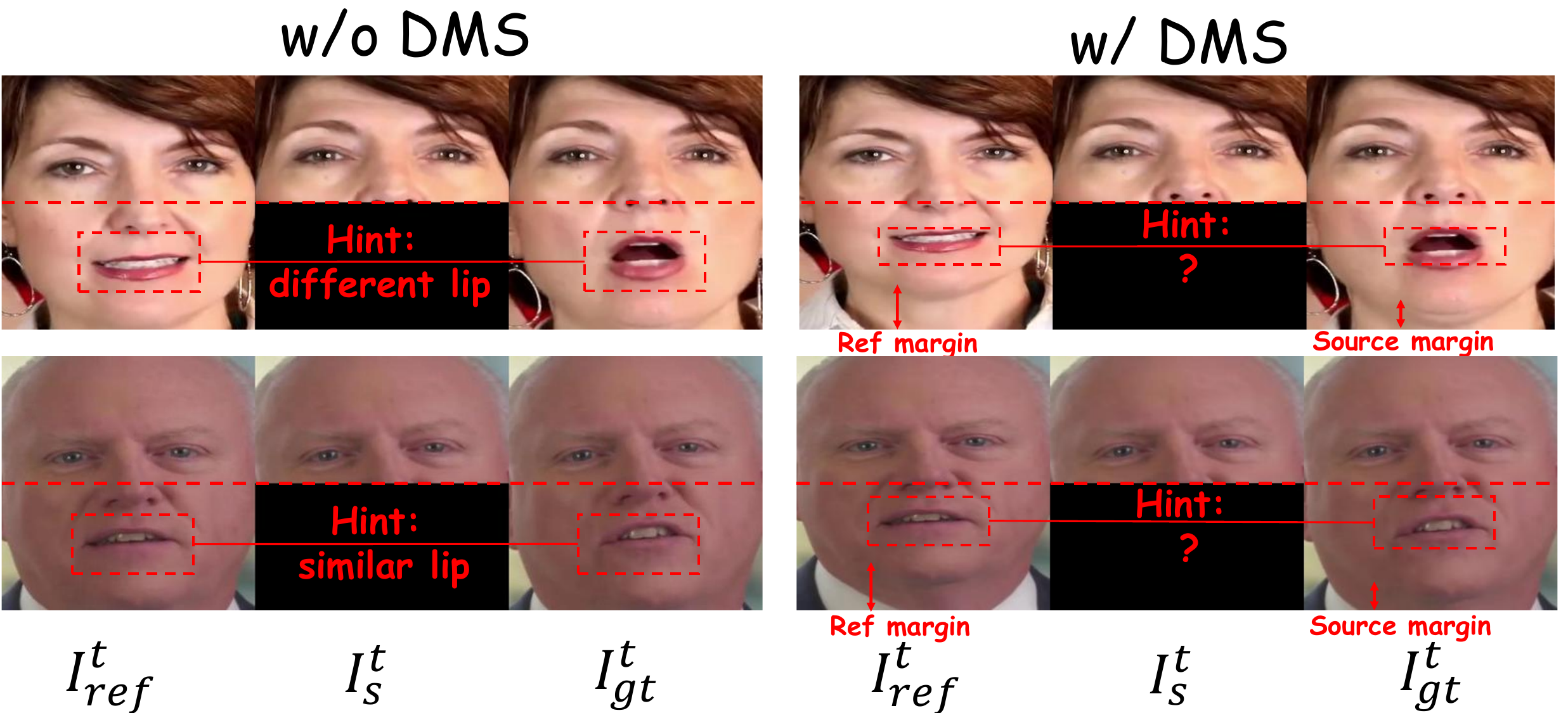}
   \caption{The principle of Dynamic Margin Sampling (DMS) in promoting lip movement learning. Without DMS, the model can easily infer the general lip shape of \(I_{gt}^{t}\) from the relative position of the nose in the input images \(I_{\text{ref}}^{t}\) and \(I_{s}^{t}\). With DMS, this cue is weakened, forcing the model to learn the lip movements.}
   \label{fig:fig_DMS}
\end{figure}

Prior works~\cite{wang2024v, jiang2024loopy, xu2024hallo} have noted that the mapping from audio to lip movements is inherently weak. Stronger conditions, such as identity constraints or other more dominant learning tasks, can easily overshadow lip-sync learning. Additionally, we have identified and localized a previously overlooked issue in previous methods~\cite{prajwal2020lip, zhong2023identity}: the leakage of lip movement information in training data pairs. 
The left side of~\cref{fig:fig_DMS} illustrates this issue.
This may lead to the model directly copying the reference's lip movements and ignoring the actual changes in lip movements, as shown in~\cref{fig:fig_loss_ab}(d).

We propose Dynamic Margin Sampling (DMS) to disrupt this implicit ``hint'' from the data.
Specially, we introduce random margins around the chin area when cropping \(I_{\text{ref}}^{t}\) and \(I_{\text{gt}}^{t}\). 
It is crucial that the margins for \(I_{\text{ref}}^{t}\) and \(I_{\text{gt}}^{t}\) are independently and randomly generated; otherwise, the hint remains. As shown in~\cref{fig:fig_DMS}, after applying DMS, the information from unmasked regions (e.g., the nose area) no longer directly indicates the degree of mouth opening, thereby forcing the model to rely on the audio input to generate accurate lip movements.

\section{Experiments}
\label{sec:experiments}

\begin{table*}[t]
\begin{center}
\begin{tabular}{lcccccccc}
\toprule
\multirow{2}{*}{\bf Method} & \multirow{2}{*}{\bf Type} & \multicolumn{3}{c}{\bf HTDF} & \multicolumn{3}{c}{\bf VFHQ} \\
\cmidrule(lr){3-5} \cmidrule(lr){6-8}
& & \multicolumn{1}{c}{FID $\downarrow$} & \multicolumn{1}{c}{CSIM $\uparrow$} & \multicolumn{1}{c}{LSE-C $\uparrow$} & \multicolumn{1}{c}{FID $\downarrow$} & \multicolumn{1}{c}{CSIM $\uparrow$} & \multicolumn{1}{c}{LSE-C $\uparrow$} \\
\midrule
Wav2Lip~\cite{prajwal2020lip} & GAN & 11.55 & 0.84 & 7.42 & 14.99 & 0.82 & 5.84 \\
VideoRetalking~\cite{cheng2022videoretalking} & GAN & 11.29 & 0.80 & 7.59 & 15.83 & 0.79 & 6.13 \\
DI-Net~\cite{zhang2023dinet} & GAN & 6.94 & 0.80 & 5.96 & 15.03 & 0.71 & 3.37 \\
IP-LAP~\cite{zhong2023identity} & GAN & 10.16 & \textbf{0.86} & 4.47 & 10.95 & \textbf{0.85} & 3.88 \\
LatentSync~\cite{li2024latentsync} & Diffusion & 8.41 & 0.84 &\textbf{7.90} & 9.89 & 0.82 & \textbf{6.79} \\
SyncLab~\cite{sync_so_projects} & – & 10.85 & \textbf{0.86} & 6.37 & 9.85 & \textbf{0.85} & 5.22 \\
\midrule
Ground Truth & – & 0.00 & 1.00 & 7.73 & 0.00 & 1.00 & 6.93 \\
\midrule
MuseTalk & GAN & \textbf{6.52} & \textbf{0.86} & 6.53 & \textbf{7.07} & \textbf{0.85} & 4.77 \\
\bottomrule
\end{tabular}%
\end{center}
\caption{Performance metrics for HDTF~\cite{zhang2021flow} and VFHQ~\cite{xie2022vfhq}.  
We omit the face restoration procedure from the original methods for fair comparison.    
SyncLab~\cite{sync_so_projects} is a commercial software with unknown technical details.
The best results are highlighted in \textbf{bold}.
}
\label{tab:performance}
\end{table*}

\subsection{Experimental Setup}
\paragraph{Model Architecture.}
MuseTalk's implementation adopts the pre-trained VAE model and the multimodal U-Net architecture from Latent Diffusion~\cite{rombach2022high}. 
For the audio encoder, we opt for the lightweight Whisper-Tiny model~\cite{radford2023robust}, which provides effective audio feature extraction. The audio features are integrated into the U-Net through cross-attention layers after undergoing reshaping and reorganization to match the required dimensions.

\paragraph{Training Details.}
We use 8 NVIDIA H20 GPUs for training.
In the Facial Abstract Pretraining stage, the model is trained with loss function \(\mathcal{L}_{\text{stage1}}\) (described in~\cref{loss:stage1}) for 200,000 steps using the batch size of 32 per GPU. 
The AdamW optimizer~\cite{loshchilov2017fixing} with a learning rate of \(2 \times 10^{-5}\) is employed. This stage costs 60 hours.

During the Lip-Sync Adversarial Finetuning stage, the model undergoes further refinement with the loss function \(\mathcal{L}_{\text{stage2}}\) (described in~\cref{loss:stage2}) for additional 20,000 steps. 
The parameter \(N\) in \(\mathcal{L}_{\text{sync}}\) (described in~\cref{loss:syncloss}) is set to 16, and the batch size per GPU is reduced to 2 to accommodate the increased computational demands of the adversarial training. The learning rate is adjusted to \(5 \times 10^{-6}\) to facilitate fine-grained updates. This finetuning stage completes in approximately 30 hours.
The loss hyper-parameters are set as follows: $\lambda_{vgg}=0.01$, $\lambda_{adv}=0.1$, $\lambda_{sync}=0.05$.

\paragraph{Dataset Preparation.}
We collected publicly available talking head datasets~\cite{zhang2021flow, xie2022vfhq}. 
To ensure high-quality data, we employed a rigorous data filtering pipeline. 
The final dataset spans approximately \textbf{24} hours in total duration. Details of the data filtering process are provided in the supplementary materials.
For evaluation, we randomly selected 26 videos from HDTF and 10 videos from VFHQ, using the remainder for training. All videos were segmented into clips for both training and testing phases. For preprocessing, we detected faces in each frame as Regions of Interest (ROIs), which were subsequently cropped and resized to \(256 \times 256\) pixels. The parameter \(k\) in the IFS was set to \(50\%\) of the video length.
The input size of discriminator \(\mathcal D_{lip}\) is set to $64\times128$.

During testing, we adopted a protocol mirroring real-world scenarios, where the video and audio inputs are sourced independently, and the reference image is extracted from the current frame. This unpaired evaluation protocol aligns with that used by Wav2Lip~\cite{prajwal2020lip} and VideoRetalking~\cite{cheng2022videoretalking}, ensuring a fair comparison.

\begin{figure*}[t]
\centering
\includegraphics[scale=0.45]{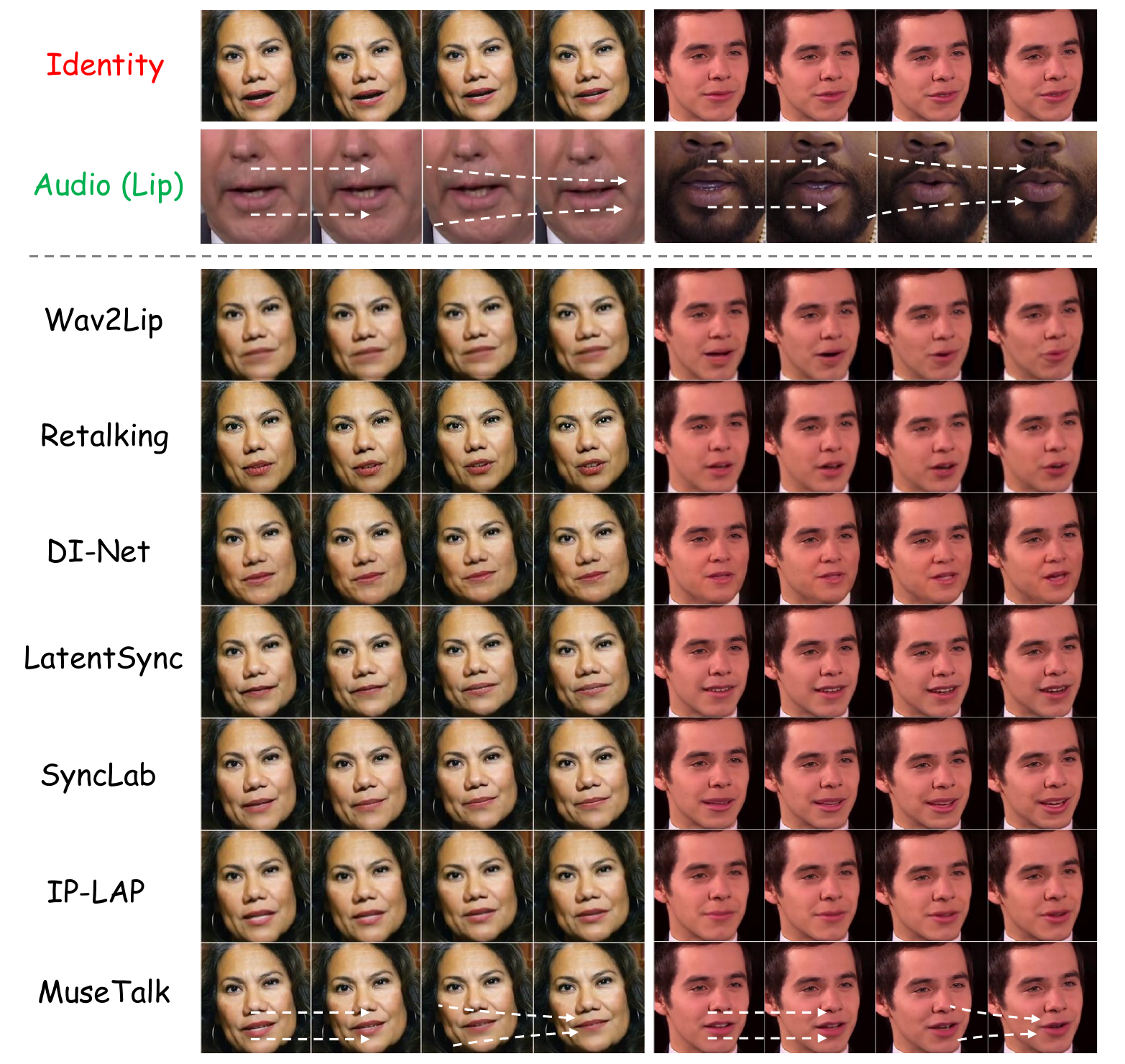}
\caption{
Qualitative comparisons on HDTF~\cite{zhang2021flow} (left) and VFHQ~\cite{xie2022vfhq} (right). 
The first two rows show the input video frames and the lips corresponding to the audio.  
The arrow shows the lip movement trend (zoom in for finer details). Additional video results are provided in the supplementary materials.
}
\label{results_sota}
\end{figure*}

\paragraph{Evaluation Metrics.}
The experiments are designed to assess the method's visual fidelity, identity preservation, and lip synchronization capabilities. 
To evaluate visual quality, we use the Frechet Inception Distance (FID)~\citep{heusel2017gans}, which measures the similarity between generated and real image distributions, providing a robust metric for visual fidelity without ground-truth talking videos.
Identity preservation is evaluated using cosine similarity (CSIM) between the identity embeddings~\cite{deng2019arcface} of the source and generated images. Lip synchronization is evaluated using lip-sync-error confidence (LSE-C)~\citep{prajwal2020lip}.

\paragraph{Compared Baselines.}
We benchmark MuseTalk against a range of SOTA video dubbing approaches.
For fair comparison, we omit the face restoration~\cite{wang2021towards} procedure if required in the original methods.
Each of them representing distinct technical advancements in the field:
(1) \textbf{Wav2Lip}~\citep{prajwal2020lip}: Pioneering work using a robust lip-sync discriminator for realistic synchronization;
(2) \textbf{VideoRetalking}~\citep{cheng2022videoretalking}: Three-stage pipeline for high-quality lip synchronization;
(3) \textbf{DI-Net}~\citep{zhang2023dinet}: Dual-encoder framework with facial action units for photo-realistic and emotion-consistent talking face videos;
(4) \textbf{IP-LAP}~\cite{zhong2023identity}: Two-stage framework combining Transformer-based landmarks with multi-reference alignment for identity preservation;
(5) \textbf{LatentSync}~\cite{li2024latentsync}: Integrates pixel-space SyncNet into latent diffusion for efficient, high-fidelity lip-sync generation;
(6) \textbf{SyncLab}~\cite{sync_so_projects}: Commercial software for lip-syncing models, focusing on cutting-edge AI video solutions.

\subsection{Quantitative Evaluation}
\paragraph{Benchmark.} 
Table~\ref{tab:performance} presents the quantitative analysis on the HDTF and VFHQ datasets. 
MuseTalk demonstrates superior performance, achieving the lowest FID scores (\textbf{6.52} on HDTF and \textbf{7.07} on VFHQ) and the highest CSIM scores (\textbf{0.86} on HDTF and \textbf{0.85} on VFHQ),  outperforming existing methods. 
While its LSE-C scores are slightly lower than some competitors, MuseTalk strikes a remarkable balance between visual fidelity and lip-synchronization accuracy.

Analyzing the baseline methods, Wav2Lip~\cite{prajwal2020lip} and VideoRetalking~\cite{cheng2022videoretalking} exhibit relatively lower visual quality, as evidenced by their higher FID scores (e.g., 11.55 vs. 6.52 on HDTF for MuseTalk). This discrepancy stems from their training on downscaled face regions (96 × 96 pixels), which compromises image clarity. In contrast, DI-Net~\cite{zhang2023dinet} achieves the second-lowest FID score on HDTF (6.94) through its deformation-based approach, which effectively preserves high-frequency texture details. However, its identity preservation capability is notably limited, as reflected in its subpar CSIM score (0.80 on HDTF and 0.71 on VFHQ). This weakness arises from its reliance on random reference image sampling, which introduces redundancy and hinders natural lip movements, ultimately affecting its LSE-C performance.

Among GAN-based approaches, IP-LAP~\cite{zhong2023identity} stands out with the highest CSIM score (\textbf{0.86}) on HDTF, showcasing exceptional identity preservation. However, it suffers from low visual quality and lip-synchronization capabilities.
Turning to diffusion models, LatentSync~\cite{li2024latentsync} achieves the best LSE-C score (\textbf{7.90} on HDTF), indicating superior lip-synchronization capabilities. However, its non-real-time nature limits its practicality for real-world applications. In contrast, MuseTalk runs in real-time, achieving 30 FPS at a 256×256 resolution on an NVIDIA V100 GPU with preloaded data.

In summary, MuseTalk is the most balanced solution, achieving great performance across multiple metrics while maintaining real-time capabilities. Its lip-sync accuracy is on par with that of the commercial software SyncLab~\cite{sync_so_projects}.

\paragraph{User Study.}
To assess the quality of lip synchronization, human judgment is relied upon. A user study was conducted to further evaluate the performance of our proposed method. For this study, dubbed videos were created by different methods using 36 unsynced audio-video pairs from the HDTF datasets and the VFHQ datasets. 
Ten participants were asked to rate each video based on visual quality, identity consistency, and lip-sync accuracy. They were provided with a five-point scale (with 1 being the lowest and 5 being the highest) for their evaluations. A total of 360 ratings were collected.

In the subjective evaluation, method names were hidden and videos were randomly shuffled to ensure unbiased assessment. Annotators saw labels like 
``Method 1'' and ``Method 2'' without knowing their specific methods, and the same label across different pairs did not correspond to the same method. This ensured fairness and prevented bias.

\begin{table}[t]
\begin{center}
\begin{tabular}{lccc}
\toprule
\multicolumn{1}{l}{\bf Method} & \multicolumn{1}{c}{\bf VQ$\uparrow$} & \multicolumn{1}{c}{\bf IC$\uparrow$} & \multicolumn{1}{c}{\bf LSQ$\uparrow$} \\
\midrule

Wav2Lip~\cite{prajwal2020lip}    & 2.19 & 3.07 & 2.70 \\
VideoRetalking~\cite{cheng2022videoretalking}   & 3.35 & 3.14 & 3.58 \\
DINet~\cite{zhang2023dinet}    & 2.92 & 2.40 & 2.57 \\
LatentSync~\cite{li2024latentsync} & 3.71 & 3.93 & \bf{4.07}\\
SyncLab~\cite{sync_so_projects} & 3.87 & 3.71 & 3.49 \\
MuseTalk  & \bf{4.26} & \bf{4.15} & 3.77 \\
\bottomrule
\end{tabular}

\caption{User Study. The best results are shown in \textbf{bold}. VQ: Visual Quality; IC: Identity Consistency; LSQ: Lip-Sync Quality.}
\label{user-study}
\end{center}
\end{table}
As indicated in~\cref{user-study}, the majority of participants awarded higher scores to MuseTalk in terms of visual quality, lip-sync quality, and identity consistency. More visualization can be found in the supplementary materials.

\subsection{Qualitative Evaluation}
Two illustrative examples are included in~\cref{results_sota}. 
Wav2Lip~\cite{prajwal2020lip} often produces synthesized mouth regions that appear blurry. 
VideoRetalking~\cite{cheng2022videoretalking} results in jagged artifacts around the lip area and overly smooths the face region. 
DI-Net~\cite{zhang2023dinet} induces noticeable changes in the subject's identity within the generated results. 
IP-LAP~\cite{zhong2023identity} maintains identity relatively well but generates inconsistent lip movements. 
LatentSync~\cite{li2024latentsync} and SyncLab~\cite{sync_so_projects} generate clear facial and dental details but require longer computation times compared to other methods. 
In contrast, MuseTalk achieves a better balance in lip movement consistency, identity preservation, and efficiency.

\subsection{Ablation Studies}
\label{subsec:exp-ab}
\paragraph{Informative Frame Sampling.}
We tested various values of \( k \) across different percentages (25\%, 50\%, and 75\%) of the total candidate frames to identify its optimal value. The results are shown in~\cref{ablation-IFS}, highlighting the effects of different \( k \) values on overall performance.

As seen in the table, the IFS strategy achieves peak performance when \( k \) is set to 50\% of the candidate frames. Specifically, the Fréchet Inception Distance (FID) reaches its minimum at 6.52, indicating the smallest discrepancy between generated and real images, and thus the highest image quality. Meanwhile, the Cosine Similarity (CSIM) reaches its maximum at 0.86, reflecting the highest similarity between generated and real images. Additionally, the LSE-C value is maximized at 6.53, further confirming the superior performance of the IFS strategy under this setting. 
In contrast, random sampling yields significantly inferior results, with an FID of 9.24, a CSIM of 0.79, and an LSE-C of 4.41. 


\begin{table}[t]
\begin{center}
\begin{tabular}{lcccc}
\toprule
\multicolumn{1}{c}{\bf Sampling strategy}  & \multicolumn{1}{c}{\bf FID$\downarrow$}  & \multicolumn{1}{c}{\bf CSIM$\uparrow$} & \multicolumn{1}{c}{\bf LSE-C$\uparrow$}  \\
\midrule
Random & 9.24 & 0.79 & 4.41 \\
IFS (k=25\%) & 8.31 & 0.83 & 2.94 \\
IFS (k=50\%)  & \textbf{6.52} & \textbf{0.86} & \textbf{6.53} \\
IFS (k=75\%)  & 11.22 & 0.72 & 3.27 \\
\bottomrule
\end{tabular}
\caption{Ablation study for sampling method on HDTF~\cite{zhang2021flow} benchmark.
The best results are shown in \textbf{bold}.}
\label{ablation-IFS}
\end{center}
\end{table}

\paragraph{Dynamic Margin Sampling.}
Dynamic Margin Sampling (DMS) generates random margins for the chin-to-boundary distance from a normal distribution \( \mathcal{N}(\mu, \sigma) \) within one standard deviation. We investigate two critical design choices: (1) the optimal margin magnitude and (2) whether to share margins between reference (\( I_{\text{ref}}^{t} \)) and source (\( I_{s}^{t} \)) images. These choices significantly affect lip generation quality.

To explore these factors, we evaluated three configurations: 
(i) a margin drawn from \( \mathcal{N}(20, 20) \) with independent margins for each image; 
(ii) a margin drawn from \( \mathcal{N}(10, 10) \) with shared margins between the reference and source images; and 
(iii) a margin drawn from \( \mathcal{N}(10, 10) \) with independent margins for each image.

\begin{table}[t]
\begin{center}
\begin{tabular}{lcccc}
\toprule
\multicolumn{1}{l}{\bf DMS setting}  & \multicolumn{1}{c}{\bf FID$\downarrow$}  & \multicolumn{1}{c}{\bf CSIM$\uparrow$} & \multicolumn{1}{c}{\bf LSE-C$\uparrow$}  \\
\midrule
 \( \mathcal N(20,20) \) + idp margin & 11.95 & 0.81 & 5.78 \\
 \( \mathcal N(10,10) \) + shared margin & \textbf{6.43} & 0.85 & 4.95 \\
 \( \mathcal N(10,10) \) + idp margin  & 6.52 & \textbf{0.86} & \textbf{6.53} \\
\bottomrule
\end{tabular}
\caption{Ablation study for different DMS settings on HDTF~\cite{zhang2021flow} benchmark.
The ``idp'' means independent.}
\label{ablation-DMS}
\end{center}
\end{table}

Table~\ref{ablation-DMS} shows that the larger margin setting (i) achieves lower FID and CSIM scores. 
This is because the highly variable margins force the model to focus more on background information, making it difficult to accurately determine the chin position and resulting in numerous artifacts in the generated images. 
The margin-sharing setting (ii) degrades lip accuracy because when the margins of \(I_{\text{ref}}^{t}\) and \(I_{s}^{t}\) are identical, the model can infer the mouth shape of \(I_{\text{gt}}^{t}\) based on the nose position, thereby reintroducing information leakage (described in~\cref{fig:fig_DMS}). 
The configuration (iii) achieves optimal performance by adopting a more moderate margin that effectively addresses information leakage without compromising FID and CSIM scores.

We use the optimal setting identified from these experiments for all evaluations described in~\cref{sec:experiments}.


\section{Conclusion}
\label{sec:conclusion}
This paper introduces MuseTalk, a novel framework for real-time, high-quality lip-synced generation for video dubbing. 
By modeling the audio-visual relationship in the VAE latent space, MuseTalk bypasses the computationally intensive diffusion process and outperforms existing state-of-the-art methods.
Its framework integrates two key innovations: Informative Frame Sampling and Dynamic Margin Sampling, which address the inherent trade-offs in GAN-based video dubbing methods, enhancing both the accuracy of lip movements and the fidelity of the generated videos. 
Comprehensive evaluations highlight MuseTalk's effectiveness, achieving the lowest FID, highest CSIM, and competitive LSE-C scores. 
MuseTalk shows promise for transforming digital communication and multimedia applications, with future work exploring multilingual support and broader virtual content creation.
{
    \small
    \bibliographystyle{ieeenat_fullname}
    \bibliography{ICCV2025-Author-Kit-Feb/main}
}

\end{document}


\maketitle

\section{Data Filtering}

We've discovered that the audio-visual consistency of the original video is crucial for model training of video-dubbing task, making data filtering an essential step. 
For instance, the VFHQ dataset contains numerous interview videos where the audio originates from someone off-camera, while the person on-screen remains silent. This type of ``dirty" data can significantly impair the effectiveness of video dubbing methods.

Our filtering approach involves calculating scores based on SyncNet, and subsequently eliminating data with low confidence and high offset. For a comparison of the datasets before and after filtering, please refer to~\cref{tab:dataset_statistics}.

\begin{table}[t]
\begin{tabular}{lccc}
\toprule
\textbf{Dataset}   & \textbf{Original Dataset}& \textbf{Filtered Dataset}\\
\midrule
HDTF~\cite{zhang2021flow}   & 15.75 & 15.32 \\
VFHQ~\cite{xie2022vfhq}   & 18.36 &  8.43 \\
\bottomrule
\end{tabular}
\caption{The variation in the duration of the dataset before and after filtering. The units in the table are measured in hours.}
\label{tab:dataset_statistics}
\end{table}

\section{Ablation Study for Multi Stage Training}
\begin{figure}[t]
\centering
\includegraphics[width=0.8\linewidth]{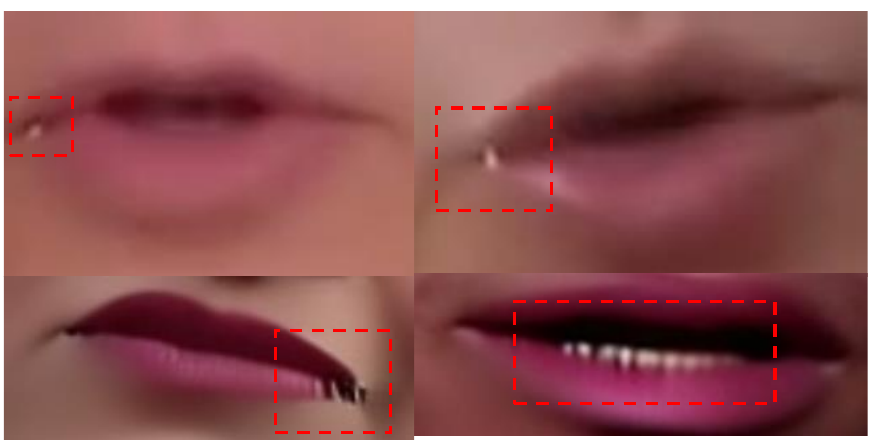}
\caption{
Artifacts observed in the generation results from single-stage training.}
\label{fig:fig_gan}
\end{figure}

As mentioned in the Section 3 in main text, optimizing \(\mathcal{L}_{\text{adv}}\) and \(\mathcal{L}_{\text{sync}}\) simultaneously in a single training stage for a less capable (randomly initialized) model leads to severe training instability. We illustrate this issue through qualitative results. As shown in~\cref{fig:fig_gan}, single-stage training introduces significant artifacts, including white spots at the corners of the mouth, missing mouth corners, and jagged teeth.

In contrast, the two-stage training strategy employed by MuseTalk first employs a more moderate training regimen in first stage to equip the model with initial face inpainting capabilities. 
This approach stabilizes the model, enabling it to achieve higher-quality results during the adversarial training in the scoond stage. 
These findings demonstrate the necessity of our proposed two-stage training strategy.

\begin{figure}[tbp]
\centering
\includegraphics[width=\linewidth]{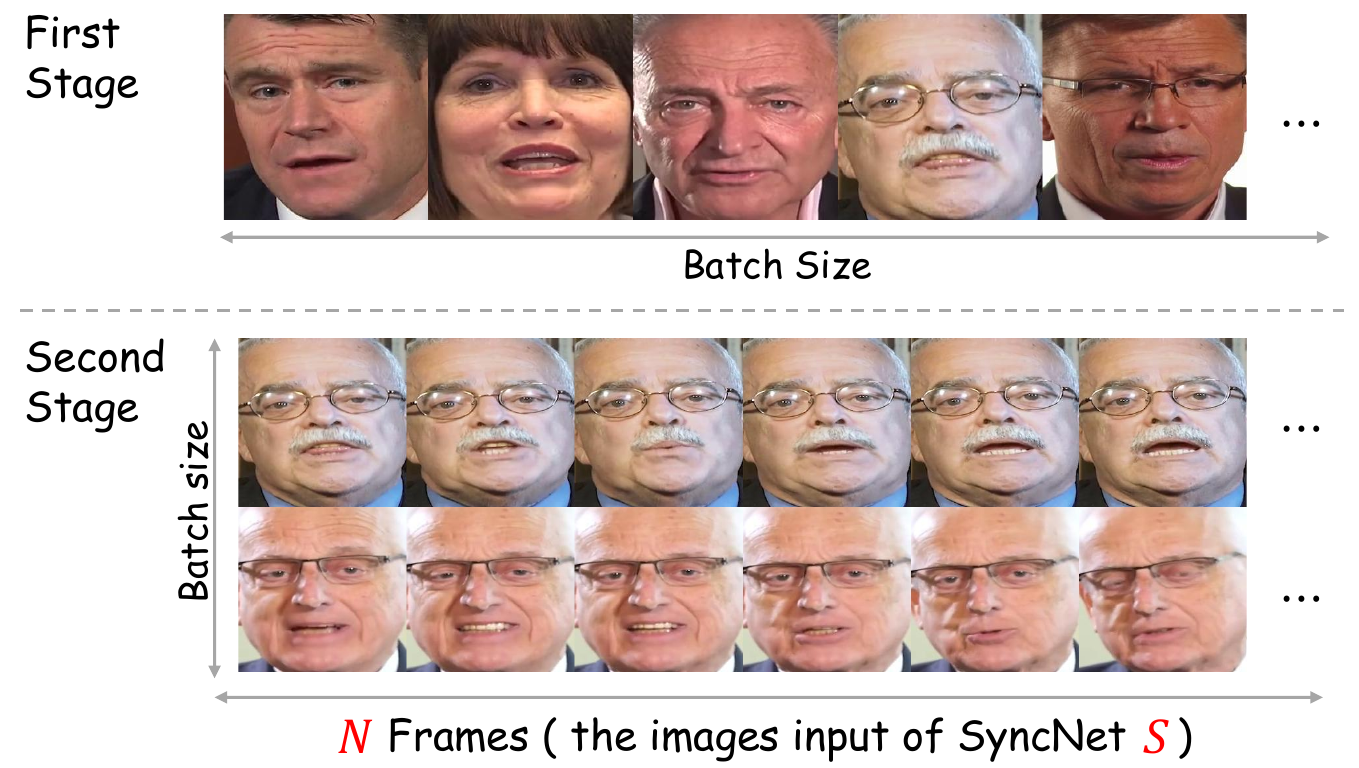}
\caption{
Data sampling strategies for different training stages.
}
\label{fig:fig_stage12}
\end{figure}

\paragraph{ID Diversity in Different Stage.}
The diagram in Fig.~\ref{fig:fig_stage12} demonstrates our application of different data sampling strategies at various training stages.

In the first stage, our goal is to expose the model to a wide array of unique identities. 
This exposure enables the model to learn a diverse range of facial details, thereby enhancing its capacity for facial abstraction.
To accomplish this, we employ a larger batch size and sample only one frame per video (identity).

In the second stage, we shift our focus towards optimizing the consistency of lip movements for a specific identity.
This stage is pivotal as it ensures the model's proficiency in accurately synchronizing lip movements with the corresponding audio, a critical aspect of video dubbing.

During the initial stage, we can configure the batch size per GPU to 32.
However, in the second stage, due to the necessity of concurrently sampling N frames to compute the sync loss (where N equals 16 in our case), the batch size per GPU is set to 2.

\section{Details for Spatial-Temporal Sampling}

\paragraph{Visualization for IFS.}

\begin{figure}[t]
\centering
\includegraphics[width=0.8\linewidth]{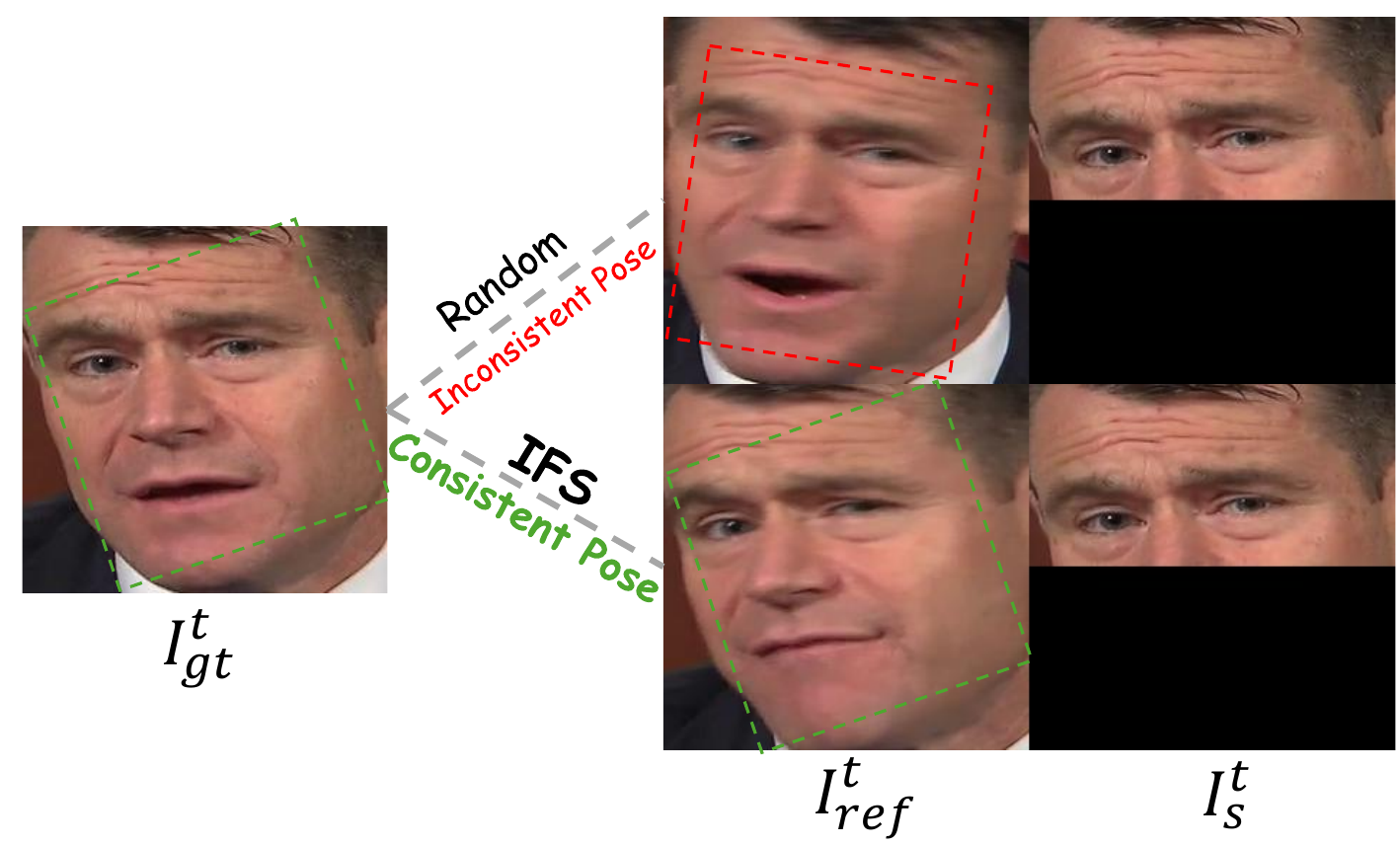}
\caption{
Comparison between random sampling and Informative Frame Sampling (IFS).}
\label{fig:fig_random}
\end{figure}
We visualize the sampled data using Informative Frame Sampling (IFS) in Fig.~\ref{fig:fig_random}. 
As shown, IFS is capable of selecting \( I^{t}_{ref} \) frames that are consistent in pose but inconsistent in lip movement with the ground-truth frame \( I^{t}_{gt} \). 
In contrast, random sampling may yield input data with large pose variations but similar lip movements compared to \( I^{t}_{gt} \).

\begin{figure*}[tbp]
\centering
\includegraphics[width=\linewidth]{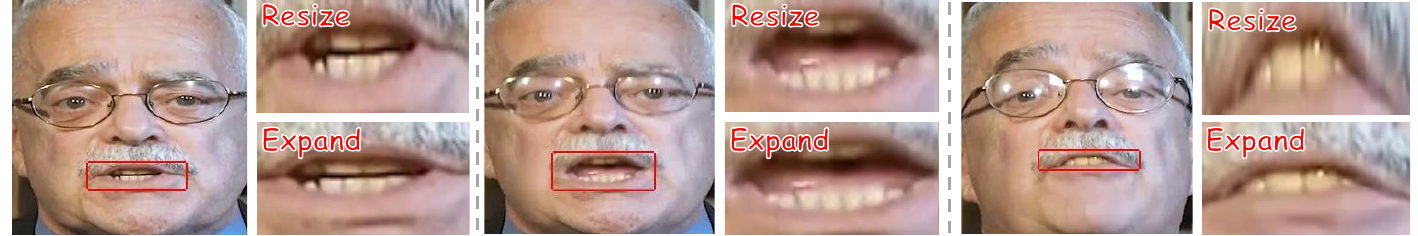}
\caption{
Visualization of the differences in processing methods for different lip regions.
}
\label{fig:fig_lip}
\end{figure*}

\section{Details for Loss Function}

\paragraph{SyncNet Loss.}
The commonly used SyncNet model for evaluating audio-driven video generation is derived from Wav2Lip~\cite{prajwal2020lip} training, which is trained on very low-resolution images ($96\times96$ pixels). 
After reviewing prior works~\cite{tian2024emo, peng2024synctalk}, we found that this model tends to produce higher lip-sync-error confidence (LSE-C)~\citep{prajwal2020lip} for low-resolution generation methods, which does not strictly reflect the lip-sync quality of the model.

Consequently, we sought a more effective audio-visual synchronization guidance model.
Recently, LatentSync~\cite{li2024latentsync} proposed a more effective SyncNet training strategy, training a SyncNet with $N = 16$ to handle $256\times256$ image inputs.
This SyncNet has approximately ten times the number of parameters compared to the model provided by Wav2Lip. Upon manual inspection, we found that this model provides more accurate audio-visual synchornize scores, which we integrated into our training pipeline.

It is crucial to note that MuseTalk's contribution lies not in providing a superior SyncNet or a more innovative loss function, but rather in harmonizing these classical losses to achieve higher-quality generation.

\paragraph{Adversarial Loss.}
During the training process, we utilize two distinct discriminators: \(\mathcal D_{face}\), which concentrates on the entire face, and \(\mathcal D_{lip}\), which specifically targets the lip region.  
The adversarial loss is incorporated using the WGAN (Wasserstein Generative Adversarial Network) framework~\cite{arjovsky2017wasserstein}, offering a more stable training experience compared to conventional GANs.
For the generator, the adversarial loss's objective is to deceive both discriminators into accepting the generated images as authentic. Specifically, the generator's goal is to minimize:
\begin{equation}
\mathcal{L}_{adv} = \mathcal{L}_{adv,face}+\mathcal{L}_{adv,lip},
\end{equation}
where 
\begin{equation}
\mathcal{L}_{adv,face} = -\mathbb{E}_{A_{mel}^{t},I^{t}_{ref}} [D_{face}(I^{t}_{o})],
\end{equation}
\begin{equation}
\mathcal{L}_{adv,lip} = -\mathbb{E}_{A_{mel}^{t},I^{t}_{ref}} [D_{lip}(I^{t}_{lip})].
\end{equation}
For the discriminators, their objectives are to differentiate between real and generated samples. 
The loss functions for the discriminators are defined as:
\begin{equation}
\mathcal{L}_{dis} = \mathcal{L}_{dis,face}+\mathcal{L}_{dis,lip},
\end{equation}

where
\begin{align}
\mathcal{L}_{dis,face} & = \mathbb{E}_{I^{t}_{real,face}}[D_{face}(I^{t}_{real,face})] 
\nonumber
\\
& + \mathbb{E}_{A_{mel}^{t},I^{t}_{ref}} [D_{face}(I^{t}_{o})]
\end{align}

\begin{align}
\mathcal{L}_{dis,lip} & = \mathbb{E}_{I^{t}_{real,lip}}[D_{lip}(I^{t}_{real,lip})] 
\nonumber
\\
& + \mathbb{E}_{A_{mel}^{t},I^{t}_{ref}} [D_{lip}(I^{t}_{lip})]
\end{align}

The discriminator operates on images of a fixed resolution. 
For \(\mathcal D_{face}\), we've matched the input region to that of MuseTalk.  
However, for \(\mathcal D_{lip}\), the lip region's size varies from frame to frame. 

We explored numerous strategies to handle this fluctuation and found that simply resizing the lip region could lead to distorted mouth shapes. This distortion could negatively impact the synchronization of audio and visuals, and compromise the overall realism.
To tackle this problem, we devised the expansion strategy. This method entails pinpointing the longer side of the mouth based on its landmarks, and then using half of this length to define the shorter side. We subsequently expand outwards to set the upper and lower breakpoints in a symmetrical manner. Following this expansion, the area is resized to a standard region, thereby safeguarding the preservation of the original mouth opening size. 
As illustrated in Fig.~\ref{fig:fig_lip}, the expansion operation guarantees that the lip region maintains its natural proportions. This results in superior mouth shape generation and improved audiovisual consistency.

\section{Blending based on Face Parsing}
\begin{figure*}[tbp]
\centering
\includegraphics[width=\linewidth]{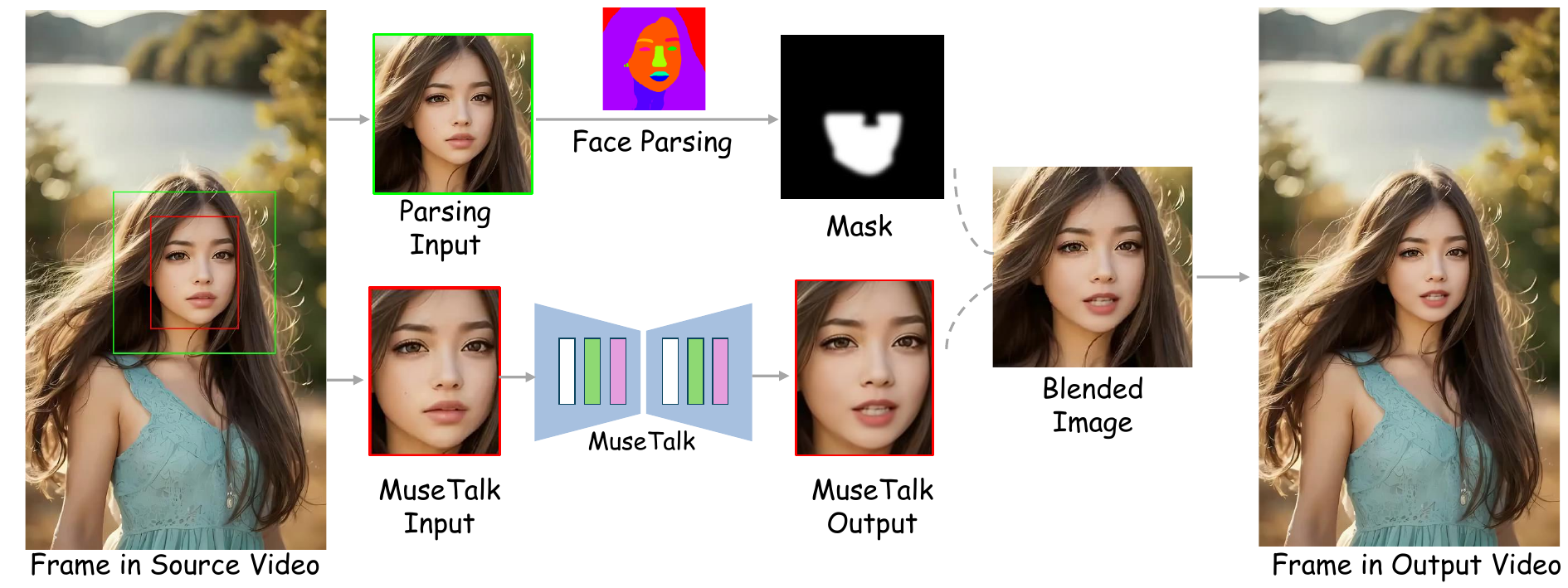}
\caption{
Illustration of the blending process used in MuseTalk.}
\label{fig:fig_blending}
\end{figure*}
We recognize that minor seams might be visible at the intersection of the generated region and the original video. To address this, we utilize a blending strategy based on face parsing.

The procedure starts by enlarging the bounding box of the identified face in the source image to create a square region. Within this area, we implement face parsing methods to generate a multi-class label map \footnote{https://github.com/zllrunning/face-parsing.PyTorch}. This map offers semantic segmentation of facial components such as eyes, mouth, nose, and so on.

Using these labels, we preserve the lower half of the face, excluding the nasal region. This choice aids in maintaining natural facial features while circumventing intricate transitions around the nose. Furthermore, we apply a Gaussian blur to the boundaries of the chosen region to ensure seamless transitions during the blending process, thereby creating a blending mask.

With this mask, we seamlessly integrate the generated region from MuseTalk with the original video content. The blending mask governs the contribution from each source.

For real-time applications, we precalculate the blending mask during the preprocessing phase as it relies solely on the original video content. This optimization considerably minimizes computational overhead during runtime, facilitating efficient processing for real-world deployment.

Figure~\ref{fig:fig_blending} depicts the entire blending process, showcasing how the proposed method delivers visually consistent results while maintaining computational efficiency.

\section{Ethical Considerations}
Technologies like MuseTalk, which facilitate video dubbing, contribute significantly to advancements in multimedia communication and entertainment. However, they also raise potential ethical concerns, particularly regarding their misuse for creating deepfakes and other forms of deceptive content. The synthesized results from MuseTalk, while highly realistic, may still display certain visual artifacts. These artifacts could serve as indicators for detecting deepfakes, thereby providing a layer of transparency and assisting in the identification of manipulated content. Furthermore, the development and deployment of such technologies should be accompanied by robust measures to prevent unauthorized use and ensure that the generated content is used ethically and legally.

In summary, while MuseTalk and similar technologies provide powerful tools for video dubbing and digital content creation, their use must be approached with caution and responsibility. It is essential to implement necessary safeguards to mitigate potential misuse and uphold ethical standards in the digital domain.

\section{Limitation and Future work}
While MuseTalk demonstrates a notable improvement in face region resolution ($256\times256$) compared to other state-of-the-art methods, it has yet to reach its full resolution potential. Additionally, certain facial details—such as mustaches, lip shape, and color—are not always well-preserved, which can affect identity consistency. Lastly, occasional jitter is observed due to the single-frame generation process, compromising smoothness.

To address these limitations, future work will focus on incorporating higher-quality training data and integrating a temporal module to reduce jitter and ensure smoother transitions. These enhancements aim to improve both resolution and overall visual consistency. Moreover, incorporating super-resolution models like GFP-GAN~\cite{wang2021towards} as a post-processing step could further elevate output quality in real-world applications.

{
    \small
    \bibliographystyle{ieeenat_fullname}
    \bibliography{ICCV2025-Author-Kit-Feb/supplement_material}
}